\documentclass[10pt,twocolumn,letterpaper]{article}

\usepackage{cvpr}              

\usepackage{graphicx}
\usepackage{amsmath}
\usepackage{amssymb}
\usepackage{booktabs}

\usepackage{enumitem}
\usepackage{verbatim}
\usepackage{pifont}
\usepackage[pagebackref,breaklinks,colorlinks]{hyperref}
\usepackage{amssymb}
\usepackage{dblfloatfix}
\usepackage{footmisc}
\usepackage{graphicx}
\usepackage{booktabs}
\usepackage{float}
\usepackage{dblfloatfix}
\usepackage[table]{xcolor}

\usepackage{multirow}

\usepackage{pifont}
\newcommand{\xmark}{\ding{55}}%
\newcommand{\cmark}{\ding{51}}%
\usepackage[pagebackref,breaklinks,colorlinks]{hyperref}

\usepackage[capitalize]{cleveref}
\crefname{section}{Sec.}{Secs.}
\Crefname{section}{Section}{Sections}
\Crefname{table}{Table}{Tables}
\crefname{table}{Tab.}{Tabs.}

\definecolor{Gray}{gray}{0.9}
\usepackage{amsmath}
\DeclareMathOperator*{\argmax}{arg\,max}

\def\cvprPaperID{31} 
\def\confName{CVPR}
\def\confYear{2023}

\begin{document}

\title{Mask-free OVIS: Open-Vocabulary Instance Segmentation \\ without  Manual Mask Annotations}
\author{Vibashan VS$^\star$\thanks{This work was done when Vibashan VS interned at Salesforce Research. Primary contact: vvishnu2@jhu.com}, 
Ning Yu$^\dagger$,
Chen Xing$^\dagger$,
Can Qin$^\ddagger$,
Mingfei Gao$^\dagger$, \\
Juan Carlos Niebles$^\dagger$, 
Vishal M. Patel$^\star$, 
Ran Xu$^\dagger$ \\
$^\star$Johns Hopkins University, $^\ddagger$Northeastern University, $^\dagger$Salesforce Research  \\
\small{\texttt{ \{vvishnu2, vpatel36\}@jhu.edu,  qin.ca@northeastern.edu,}} \\
\small{\texttt{ \{ning.yu, cxing, jniebles, ran.xu\}@salesforce.com}}
}

\maketitle

\begin{abstract}
\vspace{- 1.0 em}
Existing instance segmentation models learn task-specific information using manual mask annotations from base (training) categories. These mask annotations require tremendous human effort, limiting the scalability to annotate novel (new) categories. To alleviate this problem, Open-Vocabulary (OV) methods leverage large-scale image-caption pairs and vision-language models to learn novel categories. In summary, an OV method learns task-specific information using strong supervision from base annotations and novel category information using weak supervision from image-captions pairs. This difference between strong and weak supervision leads to overfitting on base categories, resulting in poor generalization towards novel categories. In this work, we overcome this issue by learning both base and novel categories from pseudo-mask annotations generated by the vision-language model in a weakly supervised manner using our proposed Mask-free OVIS pipeline. Our method automatically generates pseudo-mask annotations by leveraging the localization ability of a pre-trained vision-language model for objects present in image-caption pairs. The generated pseudo-mask annotations are then used to supervise an instance segmentation model, freeing the entire pipeline from any labour-expensive instance-level annotations and overfitting. Our extensive experiments show that our method trained with just pseudo-masks significantly improves the mAP scores on the MS-COCO dataset and OpenImages dataset compared to the recent state-of-the-art methods trained with manual masks. Codes and models are provided in \href{https://vibashan.github.io/ovis-web/}{https://vibashan.github.io/ovis-web/}.

\end{abstract}
\vspace{- 1.5 em}
\section{Introduction}
\label{sec:intro}
Instance segmentation is a challenging task as it requires models to detect objects in an image while also precisely segment each object at the pixel-level.
Although the rise of deep neural networks has significantly boosted the state-of-the-art instance segmentation performance\cite{he2017mask,chen2018masklab,wang2020solo}, these methods are still trained for a pre-defined set of object categories and are data-hungry \cite{o2019deep}. 
Particularly, one needs to manually annotate thousands of instance-level masks for each object category, which takes around 78 seconds per instance mask \cite{bearman2016s}. If we look at this quantitatively on Open Images \cite{kuznetsova2020open}, a large-scale dataset with 2.1M instance-level mask annotations requires around 5 years of human labour. 
Even after extensive annotation, these training datasets are still limited to a small number of categories and segmenting objects from a novel category requires further annotation. Therefore, it is difficult to scale up existing methods to segment a large number of categories due to intensive labour.

\begin{figure}[t!]
    \begin{center}
        \includegraphics[width=1.0\linewidth]{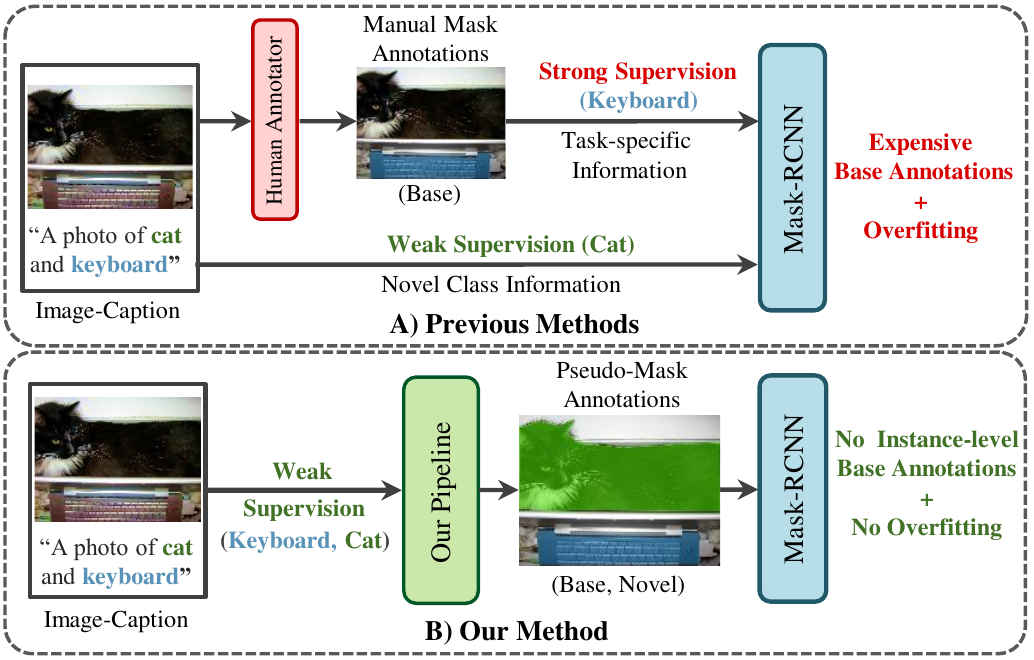}
    \end{center}
    \vskip -15.0pt
    \caption{\textbf{A) Previous Methods:} Learn task-specific information (detection/segmentation) in a fully-supervised manner and novel category information with weak supervision. During training, this difference in strong and weak supervision signals leads to overfitting and requires expensive base annotations. \textbf{B) Our method:} Given image-caption pairs, we generate pseudo-annotations for both base and novel categories under weak supervision, solving the problems of labour-expensive annotation and overfitting.  }
    \label{fig:intro} 
    \vskip -21.0pt
\end{figure}

\begin{figure*}[t!]
    \begin{center}
        \includegraphics[width=1.0\linewidth]{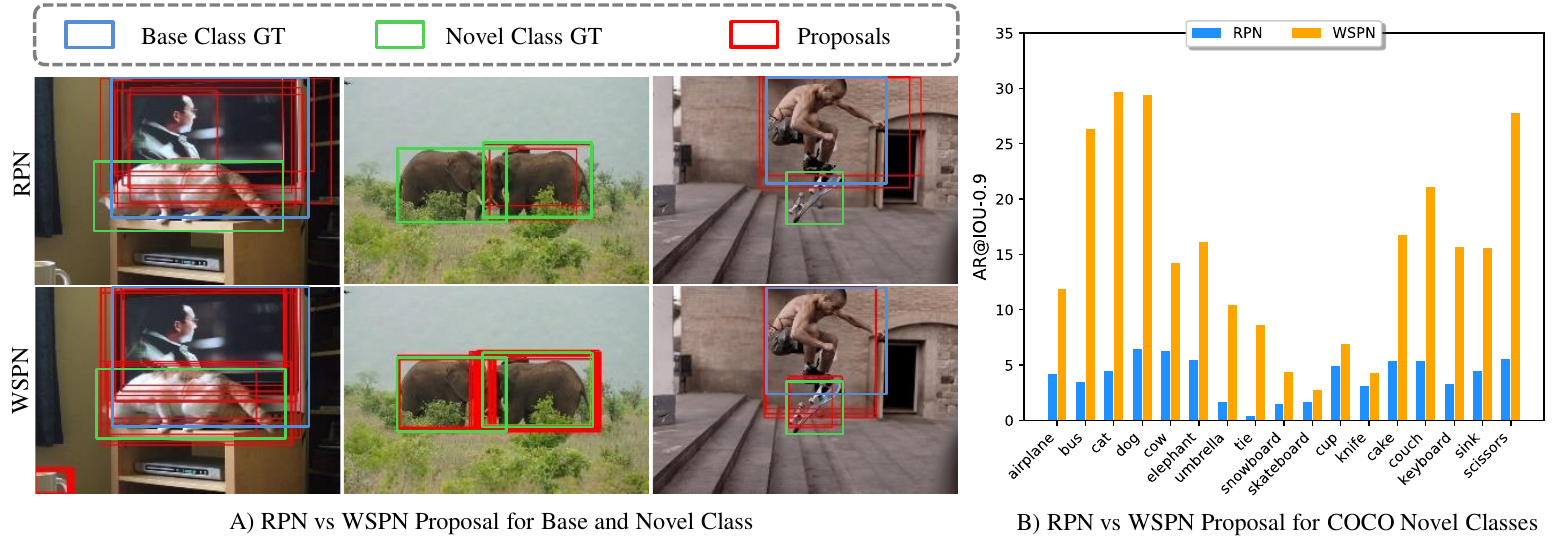}
    \end{center}
    \vskip -18.0pt
    \caption{RPN is supervised using bounding box annotations from COCO base and WSPN is supervised using image-labels from COCO base. \textbf{A)} WSPN produces better quality proposals for novel object categories compared to fully-supervised RPN. \textbf{B)} WSPN consistently produces better recall for all COCO novel categories than RPN. }
    \label{fig:motive} 
    \vskip -18.0pt
\end{figure*} 

Recently, Open-Vocabulary (OV) methods have gained much attention due to their success in detecting \cite{zareian2021open,zhong2022regionclip,gu2021open,gao2022open} and segmenting \cite{huynh2022open} \textit{novel} categories beyond \textit{base} (training) categories. 
An OV method learns task-specific information (detection/segmentation) from base categories that have manual instance-level bounding boxes or masks and learns novel category information from the pre-trained Vision-Language Model (VLM) \cite{radford2021learning,jia2021scaling} (see Fig.~\ref{fig:intro}).
All these methods produce promising results on novel categories by leveraging different forms of weak supervision such as caption pretaining \cite{zareian2021open,zhao2022exploiting}, knowledge distillation \cite{gu2021open,zhong2022regionclip} and pseudo-labelling\cite{huynh2022open,li2022grounded}. 
However, all abovementioned OV methods  still rely on the manually-annotated base categories to improve their performances on novel categories.
Without fine-tuning on base categories, existing OV methods lack task/domain specific knowledge and  the performances on novel categories will be affected~\cite{zareian2021open,gu2021open}.

Although manual instance-level annotations of base categories are critical to open-vocabulary segmentation methods, we find empirically that such fully-supervised information causes OV methods to overfit to base categories, leading to a higher failure rate when evaluated on novel categories. 
Specifically, OV methods utilize a region proposal network (RPN) \cite{ren2015faster} supervised with bounding box annotations obtained from the base categories to generate a set of bounding box proposals for all objects  categories in a given image~\cite{zareian2021open}. The feature representation from these object proposals is later matched with text embedding to learn the visual-semantic space for base and novel categories~\cite{zareian2021open}. Therefore, the quality of proposals generated for novel object categories plays a key role in determining the performance in later stages. 
However, from our experiments, we find that many objects of novel categories wouldn't be included in such proposals due to the RPN's overfitting to base categories.
Fig.~\ref{fig:motive} (A) - Top gives some examples where the RPN trained with COCO base categories fails to generate high-quality region proposals for novel categories such as elephant, cat and skateboard. Therefore, a fully-supervised proposal network is a bottleneck in OV pipeline due to its poor generalization towards novel categories. 

Given the aforementioned observations of poor generalization, we raise the question of whether we can improve the generalization by using weak supervision instead of relying on strong supervision from manual base annotations. If so, we can reduce overfitting to the base categories and the requirement for costly instance-level human annotations can be entirely removed from the pipeline. 
Our preliminary experiments give us some hope.
Our experiments show that if we train a weakly-supervised proposal network (WSPN) with image-level annotations instead of box-level annotations, the region proposals it generates can better generalize to novel objects.
As shown in Fig.~\ref{fig:motive} A), the novel objects that the RPN proposals miss are covered by WSPN. 
Fig.~\ref{fig:motive} B) shows WSPN proposals have consistently  better average recall than RPN for all COCO novel categories, indicating that WSPN proposals are more likely to cover the ground-truth bounding boxes of novel objects. 

Inspired by these observations, we propose open-vocabulary segmentation without manual mask annotations.
We do not use any human-provided box-level or pixel-level annotations during the training of our method. 
We first train a simple WSPN model with image-level annotations on base categories as a proposal generator to generate proposals for all objects given an image. 
Then, we adopt pre-trained vision-language models to select proposals as pseudo bounding boxes for novel objects. 
Given a novel object's text name, we can utilize the name as a text prompt to localize this object in an image with a pre-trained vision-language model. 
To obtain a more accurate pseudo-mask that covers the entire object, we conduct iterative masking with GradCAM\cite{selvaraju2017grad} given the vision-language model. 
Finally, we train a weakly-supervised segmentation (WSS) \cite{kim2020unsupervised} network  with previously generated bounding box and GradCAM activation map to obtain pixel-level annotation.

Our contributions are summarized as follows: \textbf{(1)}
    We propose a Mask-free OVIS pipleline where we produce manual-effort-free pseudo-mask annotations for base and novel instance segmentation using open-vocabulary and weakly supervised techniques.
\textbf{(2)} We propose a novel pseudo-mask generation pipeline leveraging a pre-trained vision-language model to generate instance-level annotations.
\textbf{(3)} Benefiting from pseudo-labels, our method sets up SOTA's for both detection and instance segmentation tasks compared to recent methods trained with manual masks on MS-COCO and OpenImages datasets.

\begin{figure*}[t!]
    \begin{center}
        \includegraphics[width=1.0\linewidth]{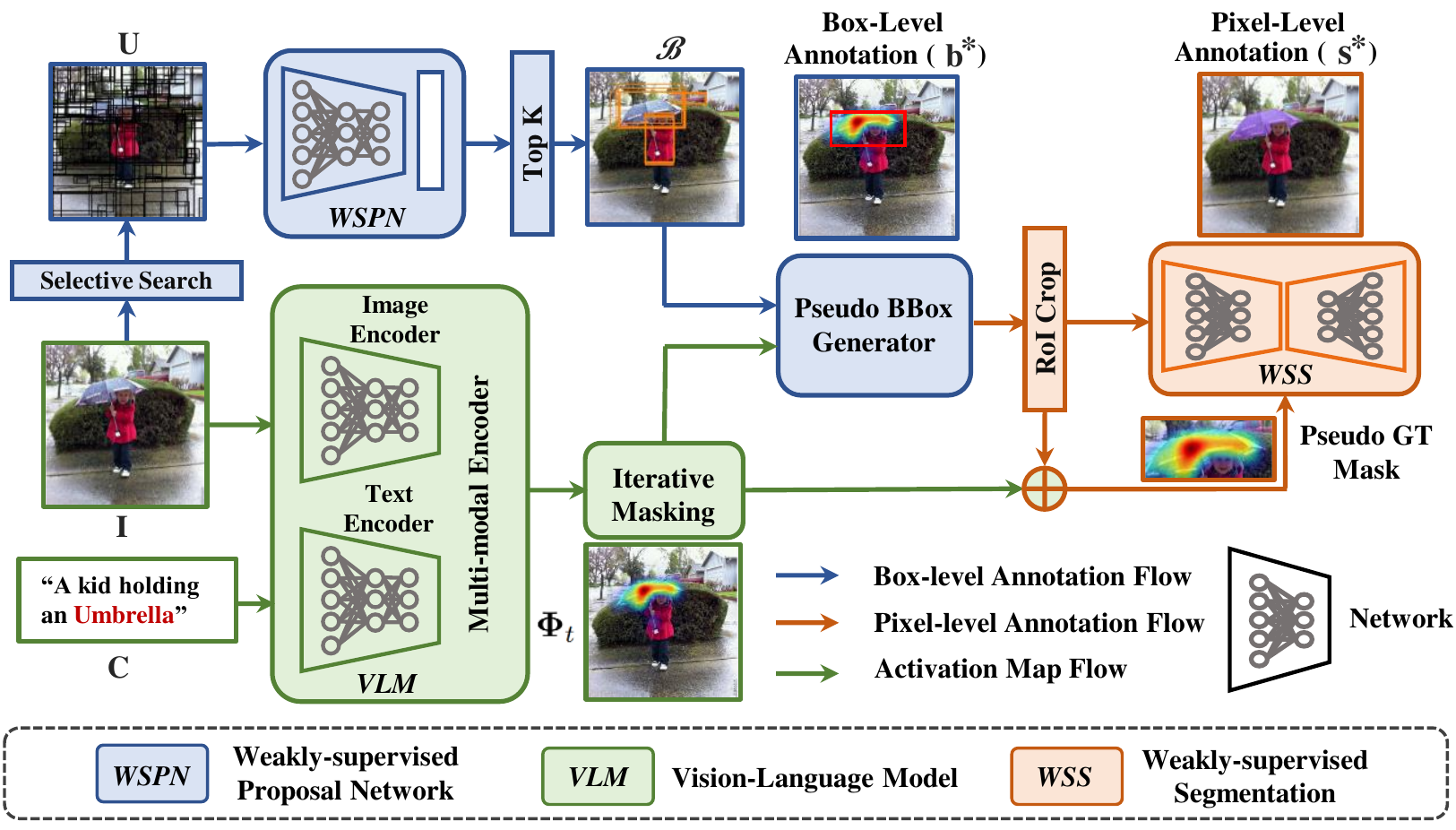}
    \end{center}
    \vspace{-15pt}
    \caption{Illustrative overview of our pseudo-mask generation pipeline. Given an image-caption pair and pre-trained VLM, we generate an activation map for the object of interest ("umbrella") and enhance it using iterative masking strategy. We generate box-level annotations using an activation map as a guidance function to select the best WSPN proposals covering the object. We crop the image corresponding to the generated pseduo bounding box and perform weakly-supervised segmentation to obtain pixel-level annotations.}
    \label{fig:archi} 
    \vspace{-15pt}
\end{figure*} 

\section{Related Work}
\noindent \textbf{Weakly-supervised learning} (WSL) methods are data efficient as they only relies on partial information about the task. They greatly reduce manual annotation effort by learning from weakly labelled data compared with fully-supervised methods. WSL applications include object detection \cite{huang2020comprehensive,vs2022instance,yang2019towards,arun2019dissimilarity,vs2023towards,zeng2019wsod2}, segmentation~\cite{lan2021discobox,tian2021boxinst,lee2021bbam}, action recognition, etc~\cite{wang2017untrimmednets,Li_2018_CVPR,zhang2021weakly}. The class activation map (CAM)~\cite{zhou2016learning} and grad-CAM~\cite{selvaraju2017grad} are popular approaches which exploits pre-trained network to generate saliency or activation map for target classes. Zhang \etal ~\cite{zhang2018adversarial} presents a CAM-based network with a classification and a localization branch for WSOD. Li \etal  ~\cite{Li_2018_CVPR} enables weakly-supervised semantic segmentation via CAM-guided attention. TS-CAM~\cite{gao2021ts} proposes a token semantic coupled attention mapping to learn the long-range visual dependency of discrete regions for object localization. 

\noindent \textbf{Vision-Language Models.}
With the rise of large-scale text-pretraining with attention-based models~\cite{devlin2018bert,zhang2022opt}, vision-language models (VLM) have caught increasing attention due to their strong performance in  downstream visual understanding tasks ~\cite{li2019visualbert,lu2019vilbert,radford2021learning}. Early VLM models~\cite{li2019visualbert,lu2019vilbert} applied the pre-trained models with constraints of the class set. With the help of contrastive learning, CLIP~\cite{radford2021learning} has enabled VLM in the wild with large-scale multimodal learning upon the 400M noisy data crawled from the web. Several methods have extended CLIP for high efficiency model training and cycle consistency~\cite{li2021align,goel2022cyclip}. BLIP~\cite{li2022blip} includes text-to-image generation as an auxiliary task so that synthetic data serves as a bonus  leading to better performance. Another method, ALBEF \cite{li2021align} utilizes cross-modal attention between image patches and text, leading to more grounded vision and language representation learning. Compared to CLIP, ALBEF excels in object localization, while CLIP performs better at zero-shot classification due to its larger training dataset.

\noindent \textbf{Open-Vocabulary.} Open-Vocabulary methods \cite{zareian2021open,rasheed2022bridging,zhao2022exploiting,bravo2022localized,ma2022open} scale up their vocabulary size for object detection or instance segmentation tasks by transferring knowledge from pre-trained vision-language models. Zareian \etal \cite{zareian2021open} proposed an open-vocabulary object detection method where a visual encoder is trained on image-caption pair to model object semantics and then transfer to zero-shot object detection. ViLD \cite{gu2021open} and RegionCLIP \cite{zhong2022regionclip} use pre-trained CLIP \cite{radford2021learning} to distill knowledge and enhance the vision-text embedding space effectively. DetPRO \cite{du2022learning} learns continuous prompt representations based on a pre-trained VLM. Similarly, Huynh \etal \cite{huynh2022open} proposes a robust pseudo-labelling, for instance segmentation where a teacher model generates pseudo-label for novel categories and is later used to supervise a student model. Therefore, all these methods require manually-annotated base categories to improve their performance on novel categories. In contrast, our method generates pseudo-labels for both base and novel categories by leveraging a pre-trained VLM, freeing the entire pipeline from human-provided instance-level annotations. Similar to our method, Gao \etal \cite{gao2022open} generates pseudo-labels using pre-trained VLM, but their approach still uses a proposal generator that relies on human-provided bounding boxes for the base categories. Our method, on the other hand, requires no such annotations, making it completely free from any human-provided box-level or pixel-level annotations.

\vspace{-2mm}
\section{Our Method}
\vspace{-2mm}
Our pipeline consists of two stages: (I) Pseudo-mask generation, and (II) Open-vocabulary instance segmentation. The goal of (I) is to generate box-level and pixel-level annotations by leveraging the region-text alignment property of a pre-trained vision-language model as illustarted in Fig. \ref{fig:archi}. In (II), we train a Mask-RCNN using the generated pseudo annotations. 
\vspace{-2mm}
\subsection{Pseudo-Mask Generation}

\noindent \textbf{Region-Text Attention Scores.} The inputs to the vision-language model are image $\mathbf{I}$ and caption $\mathbf{C} = \{c_1, c_2, ...,c_{N_{c}}\}$ pair, 
where $N_c$ is the number of words in the caption (including [CLS] and [SEP]) \cite{li2021align,devlin2018bert}. In a VLM, a text encoder is utilized to get text representations $\mathbf{T} \in \mathbb{R}^{N_C \times d}$ and an image encoder is utilized to extract region representation $\mathbf{R} \in \mathbb{R}^{N_R \times d}$, where $N_R$ is the number of regions in the image. To fuse the information from both image and text encoders, a multi-modal encoder with $M$ consecutive cross-attention layers is utilized \cite{li2021align,jia2021scaling}. These cross-attention layers learn a good region-text alignment \cite{radford2021learning,li2021align} and are utilized to obtain regions corresponding to the object of interest $c_t$ from the caption. In particular, for the $m$-th cross-attention layer, the visual region attention scores $\mathbf{X}_t^{m}$ for the object of interest $c_t$ is calculated as:
{\small
\begin{align}
    \mathbf{X}_t^{m} & =  \text{Softmax}(\frac{\mathbf{h}_t^{m-1}\mathbf{R}^T}{\sqrt{d}}), \label{eq:cross_att} \\
    \mathbf{h}_t^n & =  \mathbf{X}_t^{m}\cdot\mathbf{R}. \label{eq:cross_att_output}
\end{align}
}where  $d$ is a scalar and $\mathbf{h}_t^{m-1}$ is the hidden representation obtained from the previous $(m-1)$-th cross-attention layer.

\begin{figure}[tb]
    \begin{center}
        \includegraphics[width=1.00\linewidth]{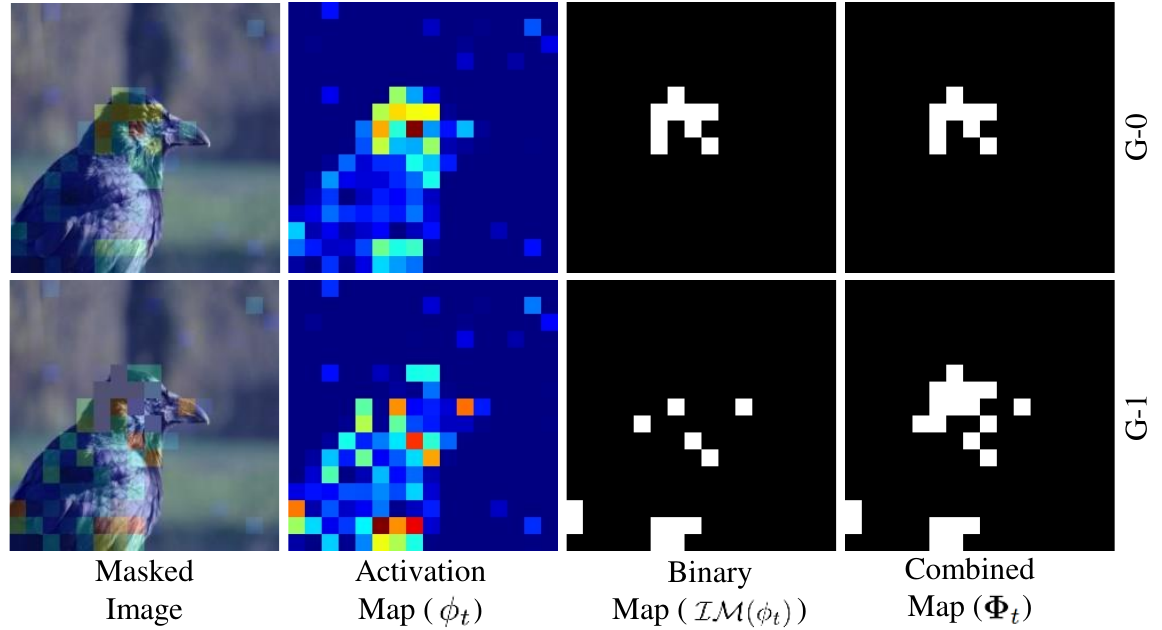}
    \end{center}
    \vspace{-18pt}
    \caption{ Comparison between activation map $\mathbf{\Phi}_t$ generated at $G$ 0 and 1. For $G$=0, the most discriminative parts of an object (bird's head) gets activated (bird's head). After masking, for $G$=1 we can observe that the activation map has shifted to less discriminative part (bird's body). Thus, by combining activation from both steps, we obtain a better activation  map trying to cover entire object $\mathbf{\Phi}_t$. }
    \label{fig:iter_vis} 
    \vskip -18.0pt
\end{figure}

\noindent \textbf{GradCAM Activation Map.}
After obtaining attention scores $\mathbf{X}_t^{m}$, we employ Grad-CAM~\cite{selvaraju2017grad} to visualize the activated regions. As in \cite{selvaraju2017grad}, we take the image-caption similarity ($S$) output from the multi-modal encoder's final layer and calculate the gradient with respect to the attention scores. The activation map $\phi_t$ for  object $c_t$ is:
{\small
\begin{equation}
    \phi_t = \mathbf{X}_t^{m} \cdot \max \left( \frac{\partial S}{\partial \mathbf{X}_t^{m}}, 0 \right). 
    \label{eq:grad}
\end{equation}
}\noindent \textbf{Iterative Masking.}
During VLM training, an object's most discriminative regions easily get aligned towards object text representation \cite{radford2021learning,li2021align}. As a result, ${\phi}_t$ is localized towards the most discriminative region and fails to cover the object completely \cite{selvaraju2017grad}. However, when we mask out the most discriminative regions, GradCAM activations are shifted towards other discriminative regions (see Fig.~ \ref{fig:iter_vis}). We propose a simple iterative masking strategy to obtain better activation where the most activated part of the object is replaced with image mean and the new activation map is computed following Eq.~\ref{eq:cross_att} and \ref{eq:grad}. The final activation map is:
\begin{equation}
    \mathbf{\Phi}_t = \bigcup_{i=1}^{G} \mathcal{IM} ({\phi}_{t}^{i})
    \label{eq:grad_final}
\end{equation}
where $G$ is a hyper-parameter indicating the number of masking iterations and $\mathcal{IM(\cdot)}$ normalize and threshold ${\phi}_t$ by 0.5. We utilize the activation map $\mathbf{\Phi}_t$ as a guidance function to generate box-level and pixel-level annotations.

\noindent \textbf{Weakly-Supervised Proposal Network.} To generate box-level annotations, we require bounding box proposals covering the activated region $\mathbf{\Phi}_t$. As explained in the introduction, the fully-supervised RPN network produces poor proposals for novel categories,  making it less preferable. To overcome this, we propose a weakly-supervised proposal network (WSPN), which generates object proposals trained in a weakly-supervised manner. Given an Image $\mathbf{I}$, the WSPN is supervised with image-labels $\mathbf{Y} = \{y_{1}, y_{2}, ..., y_{C}\}$, where $y_c$ = 0 or 1 indicates the absence or presence of class $c$ in $\mathbf{I}$ and $C$ denotes total number of classes \cite{bilen2016weakly}. First, we utilize Selective Search to generate a set of unsupervised proposal $\mathbf{U} = \{u_{1}, u_{2}, ..., u_{N}\}$ where $N$ is the total number of bounding box proposals. Then, the Image $\mathbf{I}$ and proposal $\mathbf{U}$ are fed into a CNN backbone to extract features and RoI pooling layer \cite{ren2015faster} to obtain RoI pooled feature vectors. Following \cite{bilen2016weakly}, the pooled feature vectors are passed to a classification and detection branch to generate two matrices $\mathbf{W}^{cls}, \mathbf{W}^{det} \in\mathbb{R}^{C \times N}$. Then, $\mathbf{W}^{cls}$ and $\mathbf{W}^{det}$ matrices are normalized along the category direction (column-wise) and proposal direction (row-wise) by the softmax layers $\sigma(\cdot)$ respectively. From $\mathbf{W}^{cls}$ and $\mathbf{W}^{det}$, the instance-level classification scores for object proposals are computed by the element-wise product $\mathbf{W_C} = \sigma(\mathbf{W}^{cls})\odot\ \sigma(\mathbf{W}^{det})$. and image-level classification score for the $c_{th}$ class is computed as $p_{c} = \sum_{i=1}^{ N} w_{i,c}$. Following \cite{ren2020instance}, we select the high-scoring proposals are used as pseudo-regression targets ( $\mathbf{\hat{T}} = \{\hat{t}(u_{1}), \hat{t}(u_{2}), ..., \hat{t}(u_{N})\}$) for low-scoring proposals to make sure objects can be more tightly captured. To this end, the classification loss and regression loss for WSPN are calculated as: 
{\small
\begin{equation}
\begin{gathered}
   \mathcal{L}_{wspn} = -\sum_{c=1}^{C} y_{c} \log p_{c} + (1-y_{c})\log (1-p_{c}) + \\ \frac{1}{N}\sum_{u = 1}^{N} \mathcal{L}_{smoothL1}(\hat{t}(u_i),u_i).
    \label{eq:wspn_reg_loss}
\end{gathered}
\end{equation}
}WSPN is trained to localize and classify objects by minimizing Eq.~\ref{eq:wspn_reg_loss}. The trained WSPN model is used to generate object proposals and top $K$ proposal candidates over all the classes $\mathbf{\mathcal{B}}=\{\mathbf{b}_1, \mathbf{b}_2, ..., \mathbf{b}_K\}$ are selected by sorting proposal confidence scores obtained from $\mathbf{W}^{det}$. Finally, from the top proposal candidates $\mathcal{B}$, we select the proposal which overlaps the most with $\mathbf{\Phi}_t$ as pseudo box-bounding:
\setlength{\belowdisplayskip}{0pt} \setlength{\belowdisplayshortskip}{0pt}
\setlength{\abovedisplayskip}{0pt} \setlength{\abovedisplayshortskip}{0pt}
\begin{equation}
\mathbf{b}^* = \argmax_{\mathbf{b}\in\mathcal{B}} \frac{\sum_{\mathbf{b}}\mathbf{\Phi}_t}{\sqrt{|\mathbf{b}|}},
\label{eq:box_pick}
\end{equation}
where $\mathbf{b}^*$ is the pseudo box-level annotation and  $\sum_{\mathbf{b}}\mathbf{\Phi}_t$ indicates the summation of the activation map values within a box proposal $\mathbf{b}$, and $|\mathbf{b}|$ indicates the proposal area.

\noindent \textbf{Weakly-Supervised Segmentation.}
Once we obtain the pseudo bounding box $\mathbf{b}^*$, we crop image $\mathbf{I}$ to obtain the corresponding image patch. The cropped patch is then fed into a simple three-layer CNN network to perform pixel-level segmentation. To supervise the CNN network, we generate pseudo ground-truth $\mathbf{\Theta}$  using $\mathbf{\Phi}_t$ and $\mathbf{b}^*$, where we sample $Z$ points as foreground ${F_z}=\{{f}_i\}_{i=1,.,Z}$ and background ${B_z}=\{{b}_i\}_{i=1,.,Z}$ and each point is set to 1 or 0, respectively. Specifically, the foreground and background points are sampled from the most and least activated part of $\mathbf{\Phi}_t$ inside $\mathbf{b}^*$. The pseudo ground-truth $\mathbf{\Theta}$ is of size $\mathbf{b}^*$ and we supervise the network predictions only at sampled points. Thus, the segmentation loss obtained from these weak points is computed as follows:
\begin{equation}
 \mathcal{L}_{wss} = \sum_{i=1}^{G} \mathcal{L}_{ce}(\mathbf{s}^*(f_i), \mathbf{\Theta}(f_i)) + \sum_{i=1}^{G} \mathcal{L}_{ce}(\mathbf{s}^*(b_i), \mathbf{\Theta}(b_i)),
\label{eq:seg_loss}
\end{equation}
where $\mathbf{s}^*$ is the pseudo pixel-level annotation of size $\mathbf{P}$  and $\mathcal{L}_{ce}$ indicates cross-entropy loss. 

To this end, given an image  $\mathbf{I}$ and caption  $\mathbf{C}$ pair, we generate pseudo box-level $\mathbf{b}^*$ and pixel-level $\mathbf{s}^*$ annotation for the object of interest $c_t$. In practice, the pseduo-mask annotations are generated for a pre-defined set of object categories obtained from training vocabulary. Fig.~\ref{fig:pl_vis} visualizes the generated pseduo-mask annotations for various categories. Activated regions correspond well with objects of interest and the generated pseduo-mask are of good quality.

\begin{table*}[t]
\caption{Object Detection (mAP) performances for MS-COCO under constrained and generalized setting. $\mathcal{C}_{\text{B}}$ and $\mathcal{C}_{\text{N}}$ are subset of $\mathcal{C}_{\Omega}$, where $\mathcal{C}_{\Omega}$ contains training vocabulary larger than COCO categories.} 
\centering
\vspace{-10pt}
\resizebox{.85\linewidth}{!}{\begin{tabular}{lccccccc}
\toprule
{Method} &  {\begin{tabular}[c]{@{}c@{}} Proposal \\ Generator\end{tabular}} &  {Language Supervision}         &  {\begin{tabular}[c]{@{}c@{}}  Base \\ Annotation \end{tabular}} &  {\begin{tabular}[c]{@{}c@{}}Constrained \\ Novel\end{tabular}} &  {\begin{tabular}[c]{@{}c@{}}Generalized \\ Novel\end{tabular}} \\ \hline \hline
WSDDN \cite{bilen2016weakly}                                                                    &  -   & Image-labels in  $\mathcal{C}_{\text{B}} \cup \mathcal{C}_{\text{N}}$       & \xmark                                                          & -              & 19.7        \\ 
Cap2Det  \cite{ye2019cap2det}                                                            &  -   & Image-labels in  $\mathcal{C}_{\text{B}} \cup \mathcal{C}_{\text{N}}$       & \xmark                                                          & -              & 20.3        \\ \hline 
SB   \cite{bansal2018zero}                                                             & RPN $COCO_{base}$     & -                             & \cmark                                                         & 0.70              & 0.31        \\ 
DELO  \cite{zhu2020don}                                                                & RPN $COCO_{base}$  & -                             & \cmark                                                         & 7.60              & 3.41        \\ 
PL   \cite{rahman2020improved}                                                                & RPN $COCO_{base}$  & -                             & \cmark                                                         & 10.0              & 4.12        \\ \hline
OV-RCNN  \cite{zareian2021open}                                                            & RPN $COCO_{base}$   & Image-caption in  $\mathcal{C}_{\text{B}} \cup \mathcal{C}_{\text{N}}$      & \cmark                                                         & 27.5              & 22.8        \\ 
CLIP-RPN   \cite{gu2021open}                                                          & RPN $COCO_{base}$   & CLIP image-text pair $ \mathcal{C}_{\Omega}$           & \cmark                                                         & -              & 26.3        \\ 
ViLD \cite{gu2021open}                                                               & RPN $COCO_{base}$   & CLIP image-text pair $ \mathcal{C}_{\Omega}$          & \cmark                                                         & -              & 27.6        \\ 
Detic   \cite{zhou2022detecting}                                                              & RPN $COCO_{base}$    & Image-caption in $\mathcal{C}_{\text{B}} \cup \mathcal{C}_{\text{N}}$  & \cmark                                                         & -              & 27.8        \\ 
RegionCLIP  \cite{zhong2022regionclip}                                                          & RPN $LVIS_{base}$  & Conceptual caption  $ \mathcal{C}_{\Omega}$          & \cmark                                                         & 30.8              & 26.8        \\ 

PB-OVD  \cite{gao2022open}                                                             & RCNN $COCO_{base}$    & Image-caption in  $\mathcal{C}_{\text{B}} \cup \mathcal{C}_{\text{N}}$      & \cmark                                                         & 32.3              & 30.7        \\ 

XPM   \cite{huynh2022open}                                                              & RPN $COCO_{base}$    & Image-caption in  $\mathcal{C}_{\text{B}} \cup \mathcal{C}_{\text{N}}$      & \cmark                                                         & 29.9              & 27.0        \\ 
\rowcolor{Gray} Mask-free OVIS (Ours)                                                                  & WSPN $COCO_{base}$   & Image-labels in  $\mathcal{C}_{\text{B}} \cup \mathcal{C}_{\text{N}}$       & \xmark                                                          & 31.5              & 27.4        \\ 

\rowcolor{Gray} Mask-free OVIS (Ours)                                                                 & WSPN $COCO_{base}$   & Image-labels in  $\mathcal{C}_{\text{B}} \cup \mathcal{C}_{\text{N}}$       & \cmark                                                          & \textbf{35.9}              & \textbf{31.5}        \\
\bottomrule
\end{tabular}}
\label{tab:ovd}
\vspace{-5pt}
\end{table*}

\begin{table*}[t]
    \caption{Instance Segmentation (mAP) performances for MS-COCO and Open Images under constrained and generalized setting.}
    \vspace{-10pt}
    \centering
    \resizebox{\linewidth}{!}{\begin{tabular}{lccccccc}
    \toprule
    \multirow{2}{*}{Method}  & \multirow{2}{*}{\begin{tabular}[c]{@{}c@{}}Proposal \\ Generator \\ (MS-COCO/OpenImages)\end{tabular}} & 
    \multirow{2}{*}{\begin{tabular}[c]{@{}c@{}}Base\\  Annotation\end{tabular}}  &\multicolumn{2}{c }{MS-COCO}                                                                                                                   & \multicolumn{2}{c }{Open Images}                                                                                                               \\ \cline{4-7} 
                            &                                                                                 &                                                                              & {\begin{tabular}[c]{@{}c@{}}Constrained\\ Novel\end{tabular}}  &\begin{tabular}[c]{@{}c@{}}Generalized\\ Novel\end{tabular} & {\begin{tabular}[c]{@{}c@{}}Constrained\\ Novel\end{tabular}} & \begin{tabular}[c]{@{}c@{}}Generalized\\ Novel\end{tabular} \\ \hline \hline
    OVR+OMP  \cite{biertimpel2021prior}                                           & -                                         &  \cmark                                                                 & {14.1}                                                        & 8.3                                                         & {24.9}                                                        & 16.8                                                        \\ 
    SB    \cite{bansal2018zero}                                              & -                                        &  \cmark                                                                 & {20.8}                                                        & 16.0                                                        & {24.8}                                                        & 17.3                                                        \\ 
    BA-RPN   \cite{zheng2021zero}                                         & -                                          &  \cmark                                                                 & {20.1}                                                        & 15.4                                                        & {25.3}                                                        & 16.9                                                        \\ 
    Soft-Teacher  \cite{xu2021end}                                       & RPN $COCO_{base}$/RPN $OpenImg_{base}$                                         &  \cmark                                                                 & {14.8}                                                        & 9.6                                                         & {25.9}                                                        & 17.6                                                        \\ 
    Unbiased-Teacher  \cite{liu2021unbiased}                                  & RPN $COCO_{base}$/RPN $OpenImg_{base}$                                         &  \cmark                                                                 & {15.1}                                                        & 9.8                                                         & {22.2}                                                        & 14.5                                                        \\ 
    OV-RCNN  \cite{zareian2021open}                                                   & RPN $COCO_{base}$/RPN $OpenImg_{base}$                                  &  \cmark                                                                 & {20.9}                                                        & 17.1                                                        & {23.8}                                                        & 17.5                                                        \\ 
    XPM   \cite{huynh2022open}                                             & RPN $COCO_{base}$/RPN $OpenImg_{base}$                                         &  \cmark                                                                 & {24.0}                                                        & 21.6                                                        & {31.6}                                                        & 22.7                                                        \\ 
    \rowcolor{Gray} Mask-free OVIS (Ours)                                                 & WSPN $COCO_{base}$/WSPN $COCO_{base}$                                             &  \xmark                                                                  & \textbf{27.4}                                                            &      \textbf{25.0}                                                       & \textbf{35.9}                                                             &    \textbf{25.8}                                                         \\ \bottomrule
    \end{tabular}}
    \label{tab:ovis}
    \vspace{-13pt}
\end{table*}

\subsection{ Open-Vocabulary Instance Segmentation}

After generating pseudo-mask annotations,  we train an open-vocabulary instance segmentation model. Following \cite{zareian2021open}, we employ a \texttt{Mask-RCNN} as the instance segmentation model, where a class-agnostic mask head is utilized to segment objects and the classification head is replaced with embedding head $h_{emb}$. Given a Image $\mathbf{I}$, an encoder network extracts image features and region embeddings, $\mathbf{R}=\{\mathbf{r}_i\}_{i=1,...,N_r}$, are obtained by RoI align~\cite{he2017mask} followed by a fully connected layer, where $N_r$ denotes the number of regions. The similarity between the region and text embedding pair is calculated as follows:
{\small \begin{equation}
    p(\mathbf{r}_i, \mathbf{c}_j) = \frac{\exp{(h_{emb}(\mathbf{r}_i)\cdot\mathbf{c}_j})}{\exp(h_{emb}(\mathbf{r}_i)\cdot \mathbf{bg})+ \sum_k \exp{({h}_{emb}(\mathbf{r}_i)\cdot\mathbf{c}_k})},
\end{equation}
}where, $\mathbf{C}=\{\mathbf{bg}, \mathbf{c}_1,\mathbf{c}_2,...,\mathbf{c}_{N_{C}}\}$, are object vocabulary text representation obtained from pre-trained text encoder, where $N_{C}$ is the training object vocabulary size. To learn the semantic space, negative pairs are pushed away and positive pairs are pulled together using cross entropy loss obtained from $\mathbf{b}^*$. The class-agnostic mask head is supervised by minimizing standard segmentation loss \cite{he2017mask} obtained from  $\mathbf{s}^*$.
During inference, the similarity between the region proposals embedding and text embedding from a group of object classes of interest is calculated. The region is then assigned to a class with the highest similarity. 

\section{Experiments and Results}

\noindent \textbf{Datasets.} Following \cite{zareian2021open}, we conduct experiments on MS-COCO \cite{lin2014microsoft} with data split of 48 base categories and 17 novel categories. The processed COCO dataset contains 107,761 training images and 4,836 test images. Following \cite{huynh2022open}, we conduct experiments on Open Images \cite{kuznetsova2020open} to verify the effectiveness of our method on the large-scale dataset. The Open Images dataset consists of 300 categories with a class split of  200 base categories (frequent objects)  and 100 novel categories (rare objects). Following \cite{radford2021learning}, we leverage image-labels obtained from MS-COCO and Open Images to learn the novel category information. We also experiment using image-caption datasets to show our method's effectiveness irrespective of training vocabulary.

\noindent \textbf{Evaluation Metrics and Protocols.} Following open-vocabulary methods \cite{huynh2022open,zhong2022regionclip}, for both detection and segmentation tasks, we report the mean Average Precision at intersection-over-union (IoU) of 0.5 ($mAP_{50}$). Following zero-shot settings \cite{zareian2021open}, we report novel category performance for both \textit{constrained setting} and \textit{generalized setting}. In \textit{constrained setting}, the model is evaluated only on novel class test images and in \textit{generalized setting}, the model is evaluated on both base and novel class test images.

\noindent \textbf{Implementation Details.}
In pseudo-mask generation framework, we use pre-trained ALBEF \cite{li2021align} as our vision-language model. We conducted all our pseudo-mask generation experiments using ALBEF due to the good region-text alignment when image and caption pair are present \cite{li2021align}. Following ALBEF, the cross-attention layer $m$ used for Grad-CAM visualization is set to 8. 
For attention score, we directly employ the original setting of ALBEF and no additional modification is performed. Note that other pre-trained vision-language models can also be integrated into our pipeline without major modifications. For the proposal generation pipeline, the WSPN network is trained using COCO base image-labels and the top $K$ proposals candidates is set to 50. The WSPN network is trained for 40k iterations with learn rate 0.001 and weight decay 0.0001. For iterative masking, the hyper-parameter $G$ is set to 3. In the segmentation pipeline, for each patch, the segmentation network is trained for 500 iterations with lr 0.25. 

For fair-comparison \cite{huynh2022open,zareian2021open}, we use Mask R-CNN  with a ResNet50 backbone as our open-vocabulary instance segmentation model. During pseudo-mask training, we train Mask-RCNN on MS-COCO and OpenImages using batch size 8 on 8 A5000 GPUs for 90k iterations. Following \cite{gu2021open,gao2022open}, we use text embeddings obtained from a pre-trained CLIP text encoder. During pseduo-mask training, the initial learning rate is set 0.01 and the background class weight is set to 0.2 to improve the recall of novel classes \cite{zareian2021open}. For base fine-tuning, the initial learning rate is set to 0.0005 and the weight decay is set to 0.0001. We run fine-tuning for 90k iterations where the learning rate is updated by a decreasing factor of 0.1 at 60k and 80k iterations.

\begin{table*}[t!]
\caption{Object detection and Instance segmentation ablation analysis for GradCAM, Pseudo-label and Mask-RCNN training.}
\vspace{-10pt}
\centering
\resizebox{\linewidth}{!}{\begin{tabular}{lccccccc}
\toprule
\large
\multirow{3}{*}{Method} & \multirow{3}{*}{\begin{tabular}[c]{@{}c@{}}Language \\ Supervision\end{tabular}} & \multirow{3}{*}{\begin{tabular}[c]{@{}c@{}}Mask-RCNN \\ Training\end{tabular}} &  \multirow{3}{*}{\begin{tabular}[c]{@{}c@{}}Base \\ Annotation\end{tabular}} & \multicolumn{2}{c}{{Object Detection}}                      & \multicolumn{2}{c}{{Instance Segmentation}}    \\ \cline{5-8} 
                                 &                                                                                      &                                                &                                  & { {Constrained}} &  {Generalized}    & { {Constrained}} & { {Generalized}}    \\ \cline{5-8} 
                          &      &                                                                                      &                                                                                   & { {Novel/Base}}  &  {Novel/Base/All} & { {Novel/Base}}  &  {Novel/Base/All} \\ \hline \hline
GradCAM            &  $\mathcal{C}_{\text{N}}$ &   \xmark                                                                            &       \xmark                                                                       &        8.6/0.0              &          -               &  5.2/0.0                     &             -                                 \\ 
PL            &  $\mathcal{C}_{\text{N}}$ &   \xmark                                                                            &       \xmark                                                                       &        17.3/0.0              &          -               &      14.8/0.0                 &             -                                 \\ 
PL + Mask-RCNN                      &  $\mathcal{C}_{\text{N}}$ &  \cmark                                                                      &            \xmark                                                                      &   31.1/0.4                    &  27.1/0.6/7.6   &          27.0/0.5             &   24.7/0.5/6.9                                           \\ 
PL + Mask-RCNN                     & $\mathcal{C}_{\text{B}} \cup \mathcal{C}_{\text{N}}$ &   \cmark                                                                           &       \xmark                                                                         &        32.4/22.4               &  29.3/22.8/24.5   &           27.4/18.4            &    25.0/18.3/20.1                                          \\ 
PL + Mask-RCNN                    & $\mathcal{C}_{\text{B}} \cup \mathcal{C}_{\text{N}}$ &   \cmark                                                                          &      \cmark                                                                          &      35.9/40.7                 &   40.0/31.5/37.7  &       31.2/36.7               &         36.0/28.7/34.0                                     \\
\bottomrule
\end{tabular}}
\label{tab:pl_rcnn}
 \vspace{-14pt}
\end{table*}

\subsection{Object Detection}
As shown in Table \ref{tab:ovd}, we compare our method with previous established open-vocabulary detection methods on the MS-COCO dataset. Compared to weakly-supervised methods such as WSDDN \cite{bilen2016weakly} and zero-shot methods such as SB \cite{bansal2018zero}, DELO\cite{zhu2020don}, our method outperforms them by a large margin. OV-RCNN \cite{zareian2021open}, XPM \cite{huynh2022open} are OVD methods based on caption pre-training and our method trained with only pseudo-labels improves the novel category performance by 20.2\% and 2.4\% in generalized setting, respectively. Also, when compared to the method which leverages pre-trained vision-language models such as ViLD \cite{gu2021open}, RegionCLIP \cite{zhong2022regionclip}, PB-OVD \cite{gao2022open},  our method with just pseudo-labels produces similar performance. However, with fine-tuning on base annotations, our method outperforms ViLD \cite{gu2021open}, RegionCLIP \cite{zhong2022regionclip}, PB-OVD \cite{gao2022open} by 13.3\%, 14.9\% and 2.8\% in generalized setting, respectively. This is because with fine-tuning, the model learns task/domain specific information from noise-free annotations, boosting the novel category performance. Even without base annotations, our method outperforms most of the existing OVD methods supervised using base annotations. This shows the effectiveness of our method for learning quality representation for novel categories. Specifically, the quality representation is learned due to the quality proposal generated by WSPN compared to fully-supervised RPN and RCNN proposal generators.

\subsection{Instance Segmentation}
Table \ref{tab:ovis} compares our method with previous open-vocabulary instance segmentation methods on the MS-COCO and Open Images datasets.
SB \cite{bansal2018zero} and BA-RPN \cite{zheng2021zero} are zero-shot methods which utilize different background modelling strategies and caption pre-training to learn novel categories. Compared to these, our method improves the novel category performance by a large margin on both datasets and settings. When compared against conventional pseudo-labelling methods, such as soft-teacher \cite{xu2021end} and unbiased teacher \cite{liu2021unbiased}, our method significantly improves on MS-COCO and Open Images datasets. Finally, when compared to open-vocabulary methods such as OV-RCNN \cite{zareian2021open}, XPM \cite{huynh2022open}, our method outperforms by 4.6 and 3.1 mAP in COCO and Open Images in generalized setting, respectively. All these comparisons are performed against our method trained with just pseudo-labels and no base annotation is used during training. This shows the effectiveness of our overall pseudo-mask generation pipeline.  

\subsection{Ablation Study}

\noindent \textbf{GradCAM vs Pseudo-Labels vs Mask-RCNN Training.} In Table \ref{tab:pl_rcnn}, we analyze the quality of pseudo-labels generated from the GradCAM activation map and our method (PL). After pseudo-label generation, we show how Mask-RCNN training helps to improve the quality of prediction compared to pseudo-labels. Finally, we show how fine-tuning on base annotation improves our method. In the first row, we evaluate the pseudo-mask for novel samples, where the pseudo-mask is generated by normalizing and threshold the GradCAM activation map. In the second row, we evaluate the pseudo-mask for novel samples, where the pseudo-mask is generated by our method. From Table \ref{tab:pl_rcnn}, we can observe the quality of the pseudo-mask generated by our method for novel samples is much better than the GradCAM activation map as a pseudo-mask. From Table \ref{tab:pl_rcnn} third row, we can observe training a Mask-RCNN on pseudo-labels improves the performance on novel categories by modelling fine-grained information. By including pseudo-labels from base categories, we observe that the performance on novel samples further improves. Finally,  when fine-tuning on base annotations, the performance on novel categories significantly improves by learning task/domain specific information from noise-free manual annotations.

\begin{table}[tb]
\caption{Ablation analysis for object detection and instance segmentation under different language supervision.}
\vspace{-5pt}
\centering
\resizebox{1.0\linewidth}{!}{\begin{tabular}{lccccc}
\toprule
\large
\multirow{2}{*}{Method} & \multirow{2}{*}{\begin{tabular}[c]{@{}c@{}}Language \\ Supervision\end{tabular}} & \multicolumn{2}{c}{Object Detection}                                                  & \multicolumn{2}{c}{Instance Segmentatn}                                                                                                     \\ \cline{3-6} 
       &                                                                                  & {\begin{tabular}[c]{@{}c@{}}Constrnd\\ Novel\end{tabular}} & \begin{tabular}[c]{@{}c@{}}Genrlzd\\ Novel\end{tabular} & {\begin{tabular}[c]{@{}c@{}}Constrnd\\ Novel\end{tabular}} & \begin{tabular}[c]{@{}c@{}}Genrlzd\\ Novel\end{tabular} \\ \hline \hline
{\small OV-RCNN     \cite{zareian2021open}}                                                                     &  Image-caption COCO                                                                     & {27.5}                                                        & 22.8                                                        & -                                                        & -                                                        \\ 
PB-OVD   \cite{gao2022open}                                                                                   &  Image-caption  COCO                                                                     & -                                                        &                       29.1                                  & -                                                        & -                                                        \\ 
PB-OVD   \cite{gao2022open}                                                                                   &  {\small Img-Cap  COCO, SBU, VG}                                                                     & 32.3                                                        & 30.8                                                        & -                                                        & -                                                        \\ 
Ours  &  Image-labels COCO  
        & 35.9                                        & 31.5                                        & 31.0                                        &  28.3                                                       \\ 
Ours    &  Image-caption  COCO                             & 36.1                                                & 31.8                                                & 31.5                                                & 28.8                                                        \\ \bottomrule
\end{tabular}}
\label{tab:lang_ab}
\vspace{-10pt}
\end{table}

\begin{figure*}[tb]
    \begin{center}
        \includegraphics[width=1.0\linewidth]{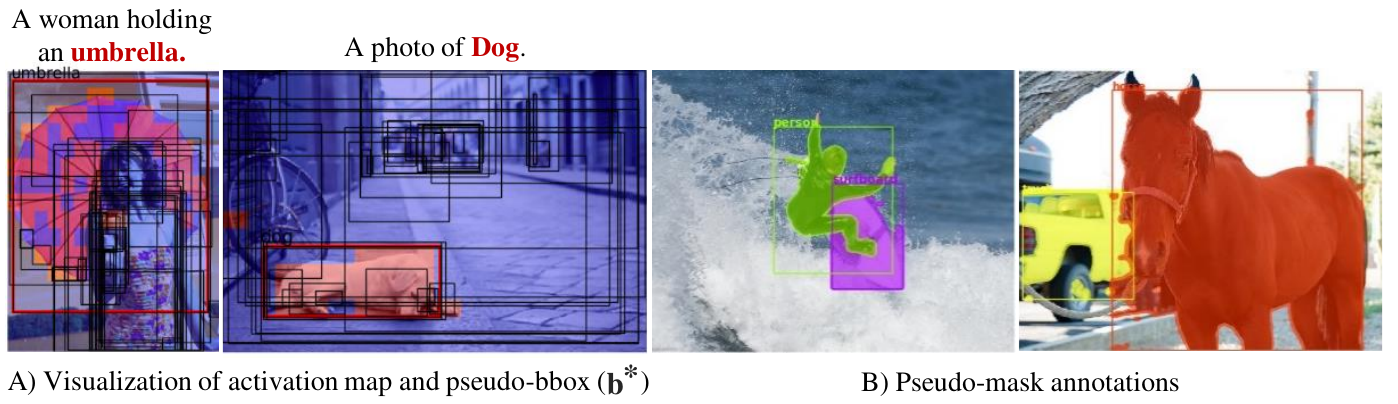}
    \end{center}
    \vspace{-18pt}
    \caption{A) Pseduo bounding box selection guided by GradCAM activation. B) Visualization of pseudo-mask generated for Open Images. }
    \label{fig:pl_vis} 
    \vskip -12.0pt
\end{figure*} 

\noindent \textbf{Language supervision.} Given an image and caption pair, our method can generate a pseudo-mask leveraging a pre-trained vision-language model. Thus to analyze the effect of captions, we conduct experiments between human-provided captions and pseudo-captions generated from image-labels. As show in Table \ref{tab:lang_ab}, human-provided captions and image-labels based pseudo-caption produce similar performance showing that irrespective of caption type, our method can generate pseudo-mask for the object of interest (see supplementary material for visual comparison). Therefore, our method is more data efficient as it requires cheap image-labels compared to human-provided captions. Table \ref{tab:lang_ab} compares our method with other pseudo-label generation methods trained with extra language supervision \cite{gao2022open}. Our method, with lesser language supervision, outperforms \cite{gao2022open} by a considerable margin. 

\begin{table}[tb]
\caption{Ablation analysis for object detection under different proposal generator. All models are fined-tuned on COCO base.}
\vspace{-8pt}
\centering
\resizebox{0.9\linewidth}{!}{\begin{tabular}{lccc}
\toprule
Method  & {\small Proposal Generator} & \begin{tabular}[c]{@{}c@{}}{\small Constrained}\\ Novel\end{tabular} & \begin{tabular}[c]{@{}c@{}}{\small Generalized}\\ Novel\end{tabular} \\ \hline \hline
{\small OV-RCNN  \cite{zareian2021open}} & {\small RPN COCO Base}      & 27.8                                                        & 22.8                                                        \\ 
{\small PB-OVD \cite{gao2022open}} & {\small RCNN COCO Base}     & 32.3                                                        &   30.8                                                      \\ 
Ours    & {\small Selective Search}   &   34.5                                                      &    31.0                                                     \\ 
Ours    & {\small WSPN COCO Base}     &   35.9                                                      &     31.5                                                    \\ \bottomrule
\end{tabular}}
\label{tab:prop_ab}
 \vspace{-14pt}
\end{table}

\noindent \textbf{Proposal Generator Quality vs Performance.} In general, better quality of proposal provides better quality of pseudo-labels. Therefore, we generated pseduo-labels using different proposal generator and the results are reported in  Table \ref{tab:prop_ab}. As shown in Table \ref{tab:prop_ab}, our method trained with WSPN as proposal generator is produces better performance compared to methods which rely on fully-supervised proposal generator such as RPN and RCNN. Also when compared to selective search as proposal generator, WSPN demonstrates better performance for novel categories. This is because WSPN refines selective search proposals  and localizes them towards objects producing better quality proposals.   

\begin{figure}[tb]
\vspace{-5pt}
\begin{center}
\includegraphics[width=0.49\linewidth]{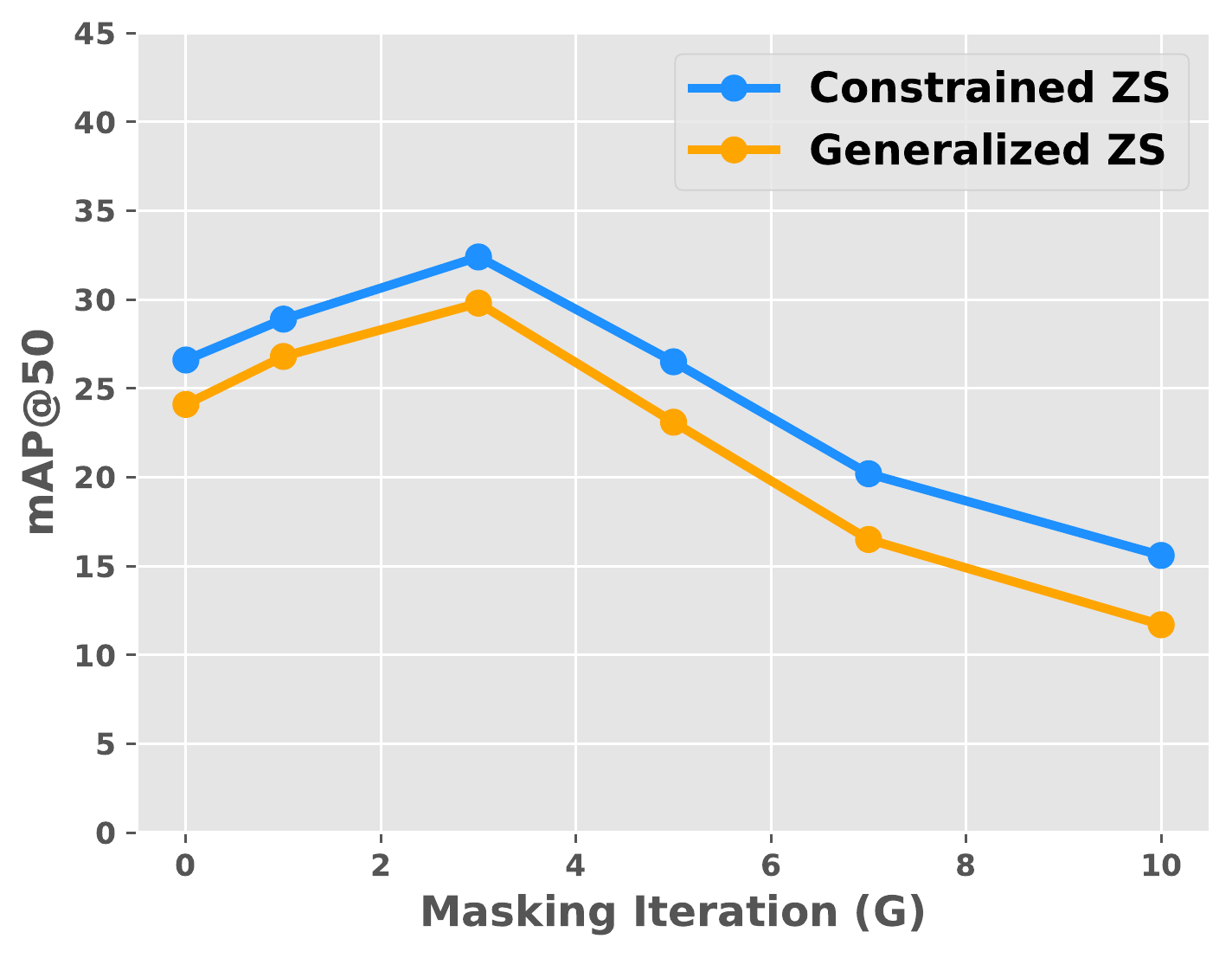}
\includegraphics[width=0.49\linewidth]{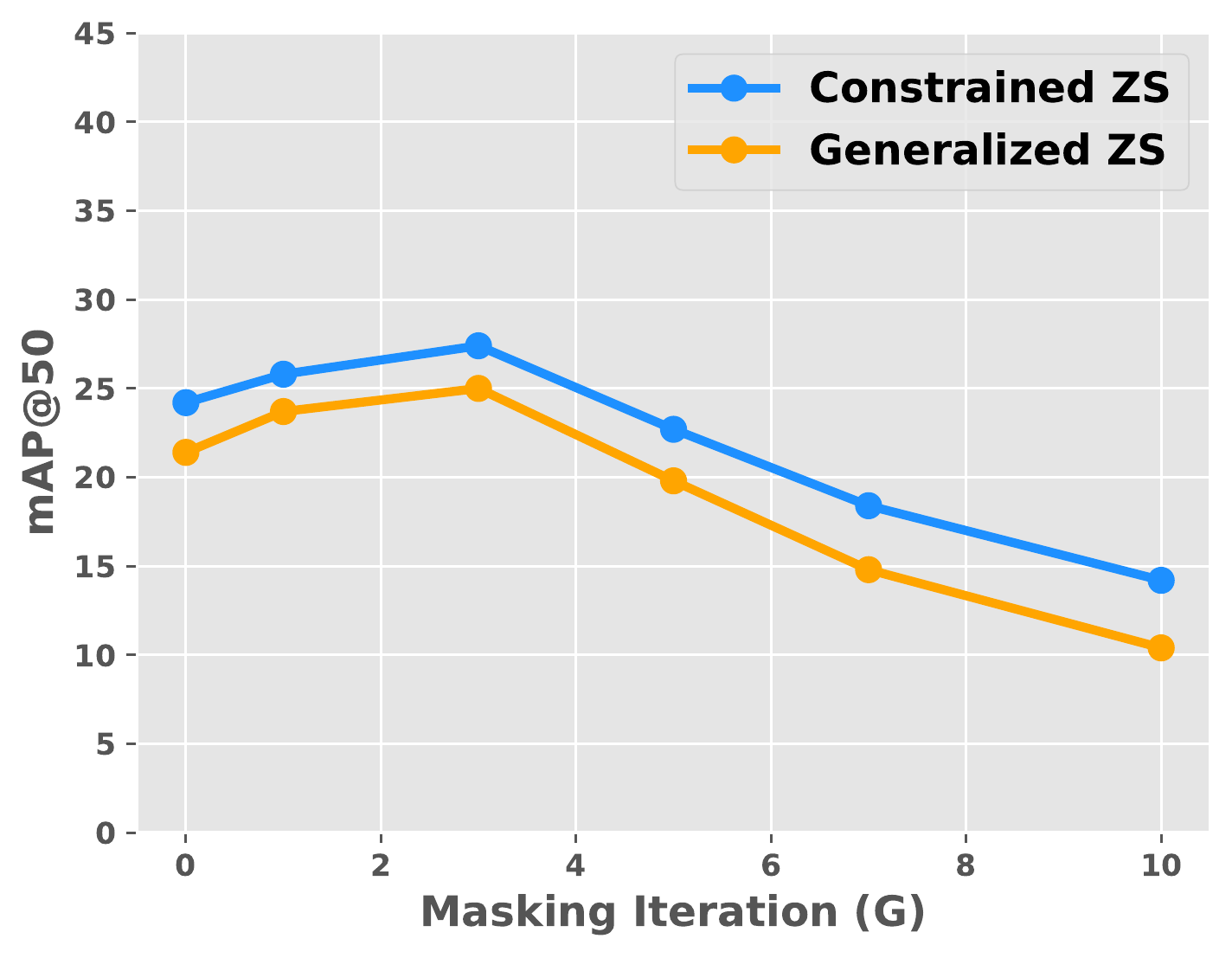}
\vskip -1pt
\hskip 10pt \footnotesize{(a) Object detection} \hskip 45pt  \footnotesize{(b) Instance segmentation}
\end{center}
 \vspace{-10pt}
 \caption{Ablation study for object detection and instance segmentation by increase number of masking iteration during PL generation. As iteration increases, GradCAM activation becomes noisy resulting in poor guidance and pseudo-mask generation.}
\label{fig:iter_mask}
 \vspace{-18pt}
\end{figure}

\noindent \textbf{Iterative Masking Steps vs Performance.} Fig. \ref{fig:iter_mask} presents ablation analysis for iterative masking hyper-parametr $G$. We see that for initial few masking steps the performance increase. Because, without any masking the GradCAM activation map is towards most discriminative parts of an object. As we mask out these region, the less discriminative regions gets activated and combining activation from previous steps produce a strong guidance function $\mathbf{\Phi}_t$. However, performing too many mask iterations could completely mask out the object of interest and unrelated background regions might get activated resulting in a poor $\mathbf{\Phi}_t$.

\noindent \textbf{Weakly-supervised Instance Segmentation.}  
Weakly-supervised Instance Segmentation (WSIS) methods performs instance-segmentation using supervision from image-labels. Similar to WSIS setting, our method performs instance-segmentation using just image-labels. Following this, we simply extended our method to WSIS setting by comparing with VOC benchmark. PENet \cite{ge2019label} is a recent WSIS method, which relies on multiple downstream task network such as segmentation, detection and multi-class classification to perfrom segmentation. CL \cite{hwang2021weakly} and IAM \cite{zhu2019learning} are WSIS methods utilize advanced attention mechanism to generate instance segmentation mask. As shown in Table \ref{tab:wsis}, our method outperform existing WSIS methods by considerable margin. Thus, the proposed method is simple and can be easily extended to OV setting or WSIS setting producing SOTA performance. Note that, other OV methods cannot be extended to WSIS setting as they rely on base annotation and similarly WSIS methods cannot be extended to OV setting due to constrained vocabulary space.

\begin{table}[tb]
\caption{Instance segmentation(mAP) performance for VOC in weakly-supervised instance segmentation setting.}
\centering
\vspace{-8pt}
\resizebox{0.65\linewidth}{!}{\begin{tabular}{lcc}
\toprule
Method & {Supervision} & {mAP}              \\ \hline \hline
Mask-RCNN  \cite{he2017mask}     & F           & 67.9                      \\ \hline
OCIS \cite{cholakkal2019object}           & I           & 26.8                      \\ 
PRM  \cite{zhou2018weakly}           & I           & 28.3                      \\ 
IAM  \cite{zhu2019learning}           & I           & 30.2                      \\ 
CL  \cite{hwang2021weakly}            & I+C         & 30.2                      \\ 
PENet  \cite{ge2019label}         & I           & 38.1                      \\ 
\rowcolor{Gray} Ours       & I           & \textbf{38.6}                      \\ \bottomrule
\end{tabular}}
\label{tab:wsis}
 \vspace{-15pt}
\end{table}

\noindent \textbf{Qualitative Analysis.} 
Fig. \ref{fig:pl_vis} (A) presents a visualization of activation maps generated for the object of interest (woman and dog). As we can see, the generated activation map covers the entire object and it can be used as a guidance function to choose the best bounding box proposal. Note that the activation are square-shaped because the original activation map is 1/16'th of the image size. We perform nearest interpolation to obtain an activation map of image size. In Fig  \ref{fig:pl_vis} (B), we visualize the pseudo-mask for Open Images generated from our pipeline. We can observe that the generated pseudo-mask is of good quality; still contains some false positives. However, with Mask-RCNN training, the model learns to filter the noise present in the pseudo-mask producing better-quality predictions.

\section{Conclusion}

We propose a novel pipeline called  Mask-free OVIS for open-vocabulary instance segmentation, which is free from any human-provided instance-level annotations. We achieve this by generating pseudo-mask annotation for base and novel categories by leveraging a pre-trained vision-language model and weakly supervised techniques. The generated pseudo-mask annotations are then used to supervise the Mask-RCNN model. Extensive experiments are conducted to show the effectiveness of our method on MS-COCO and Open Images datasets. Instead of relying on labour-expensive instance-level annotations, we hope our simple and effective pipeline provides an alternative to detect and segment long-tailed novel categories.

{\small
\bibliographystyle{ieee_fullname}
\bibliography{egbib}
}

\newpage

\onecolumn

\begin{center}
 {\large Supplementary material for Mask-free OVIS: Open-Vocabulary \\ Instance  Segmentation  without  Manual Mask Annotations}
\end{center}
    

\begin{figure*}[thb]
\begin{center}
\includegraphics[width=0.9\linewidth]{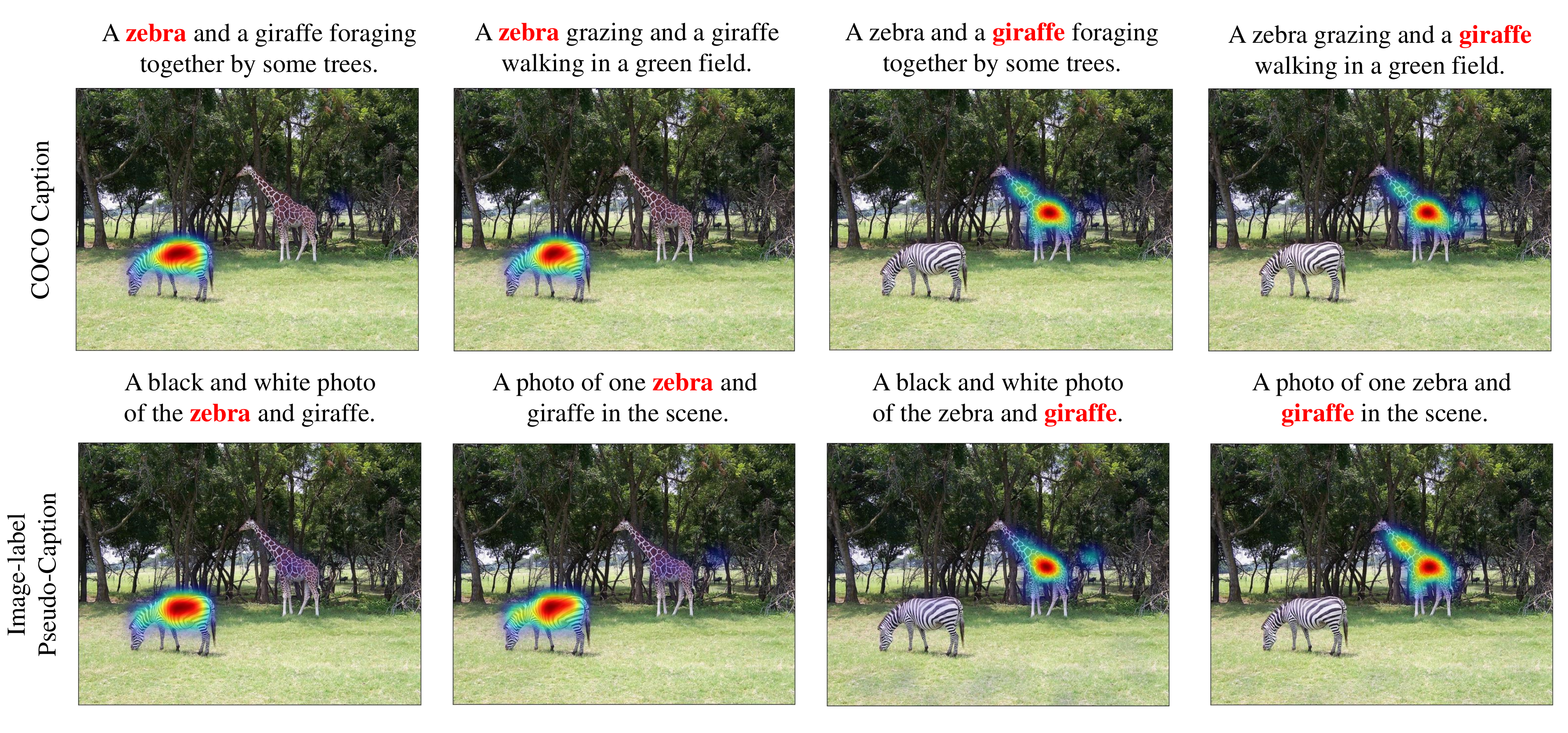}
\end{center}
\vskip -10.0pt \caption{\textbf{Top row:} Given COCO caption and object of interest $c_t$ ("zebra","giraffe"), corresponding activation map is generated using GradCAM. \textbf{Bottom row:} Given pseudo-caption generated from image-labels and object of interest $c_t$ ("zebra","giraffe"), corresponding activation map is generated using GradCAM. The original activation map are of 1/16'th of the image size and we perform bilinear interpolation to obtain an activation map of image size.} 
\label{fig:cap_vs_il}
\end{figure*}

\noindent \textbf{COCO Caption vs Image-label pseudo-caption:} \\
\noindent \textbf{Pseudo-caption generation:} Since the pre-trained vision-language models are trained on full sentences, we need to feed the image-labels into a prompt template first, and use them to generate a pseudo-captions. Specifically, given image-labels [category-1,category-2...,category-n], we randomly sample a prompt from 63 prompt templates \cite{radford2021learning, gu2021open} and the pseudo-caption are generated as "\{Prompt-x\} + \{category-1 and category-2 and ... category-n\}". For example, as shown in Fig. \ref{fig:cap_vs_il} bottom row - the sampled prompts are \textit{"A black and white photo of the \{category\}."} and \textit{"A photo of \{category\} in the scene."} and the image-labels are "zebra" and "giraffe". Thus, the generated pseudo-captions are \textit{"A black and white photo of the zebra and giraffe."} and \textit{"A photo of one zebra and giraffe in the scene."}
\\

\noindent \textbf{COCO Caption and Pseduo-caption activation map:} Given a caption and object of interest $c_t$ ("zebra","giraffe"), corresponding activation map is generated using GradCAM \cite{selvaraju2017grad} and pre-trained vision-language model \cite{li2021align}. As we can observe from Fig. \ref{fig:cap_vs_il}, both human-provided COCO captions and image-labels based pseudo-captions produce similar activations for "zebra" and "giraffe". Even for the same image with different human-provided COCO captions, the generated activation maps are similar for "zebra" and "giraffe". This shows that irrespective of caption type, the GradCAM generates similar activation map for the object of interest resulting in similar pseudo-masks.

\noindent \textbf{Generalization of WSPN network trained on VOC (20 categories) and test on COCO (80 categories) dataset:} 
\begin{figure*}[b]
\vskip -20.0pt
\begin{center}
\includegraphics[width=0.84\linewidth]{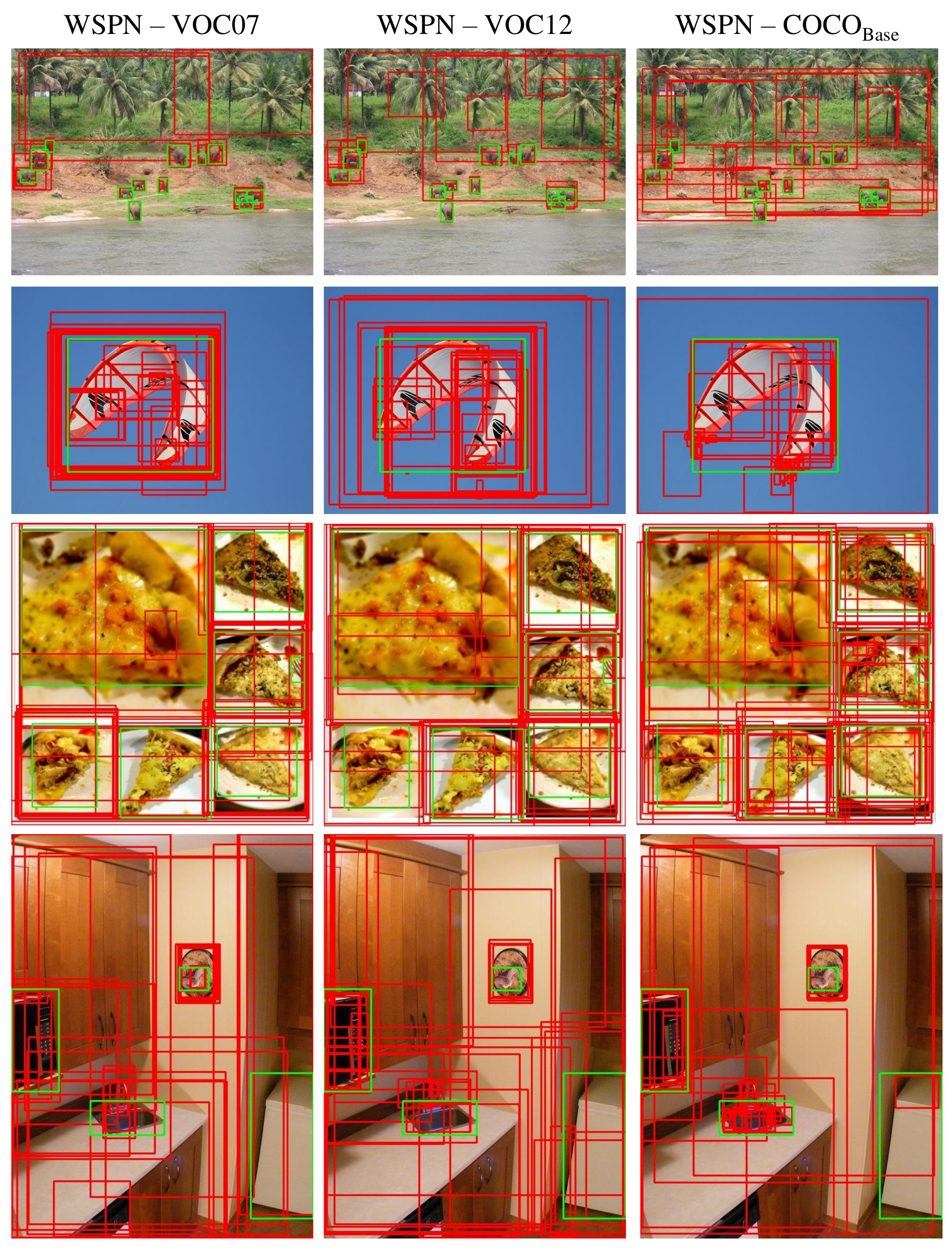}
\end{center}
\vskip -20.0pt \caption{Qualitative analysis of WSPN network trained on VOC 2007 (20 categories), VOC 2012 (20 categories) and COCO Base 2017 (48 categories) using image-labels in a weakly-supervised manner and tested on COCO 2017 (80 categories). Here, the WSPN trained on VOC 2007 and VOC 2012 has not seen categories such as \textit{"Pizza", "Kite", "Clock"} etc during training. Still, we can observe the VOC trained WSPN network can localize these unseen categories similar when tested on COCO 2017. This is due to weakly-supervised training, where the WSPN network learns to localize the object irrespective of categories it is trained on. As a result, the WSPN network trained on VOC is able to localize COCO object categories which are not seen during training. This show the generalization capability of the WSPN model and can be used to localize any objects for different dataset. {\color{green}Green:} Ground truth bounding box, {\color{red}Red:} Top 50 WSPN proposals. }
\label{fig:wspn_ab}
\end{figure*}

\begin{figure*}[t!]
\begin{center}
\includegraphics[width=0.85\linewidth]{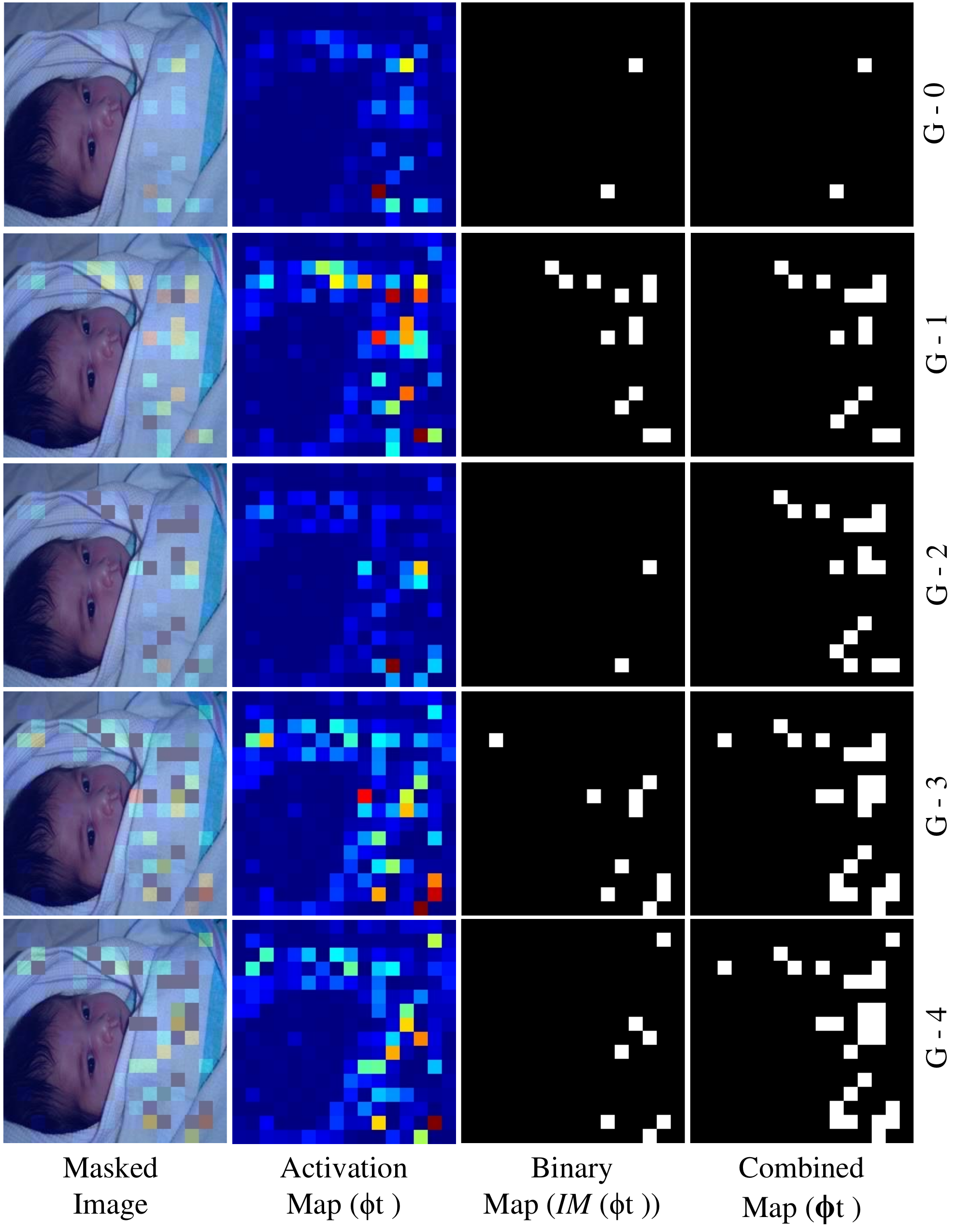}
\end{center}
\vskip -20.0pt \caption{Iterative masking visualization for "Towel" category.}
\label{fig:itermask1}
\vskip -20.0pt 
\end{figure*}
\noindent \textbf{Iterative masking lead to better guidance function:} \\
Fig. \ref{fig:itermask1} present iterative masking visualization for $G$ going from 0 to 4.
From Fig. \ref{fig:itermask1}, we can observe that, the initial GradCAM~\cite{selvaraju2017grad} activation for the towel category is less at $G$=0. Utilizing this activation map as guidance function to generate pseudo-labels will generate less accurate pseudo-labels. However, by performing the proposed iterative masking strategy, we can observe that the activation map is progressively shifting towards a less discriminative part in successive iterations and the combination of all activation's $\mathbf{\Phi}_t$ covers the entire object effectively. Note that even after multiple iterations, there is no GradCAM~\cite{selvaraju2017grad} activation for the baby, which shows the robustness of the pre-trained model towards region-text alignment.

\begin{figure*}[t!]
\begin{center}
\includegraphics[width=0.85\linewidth]{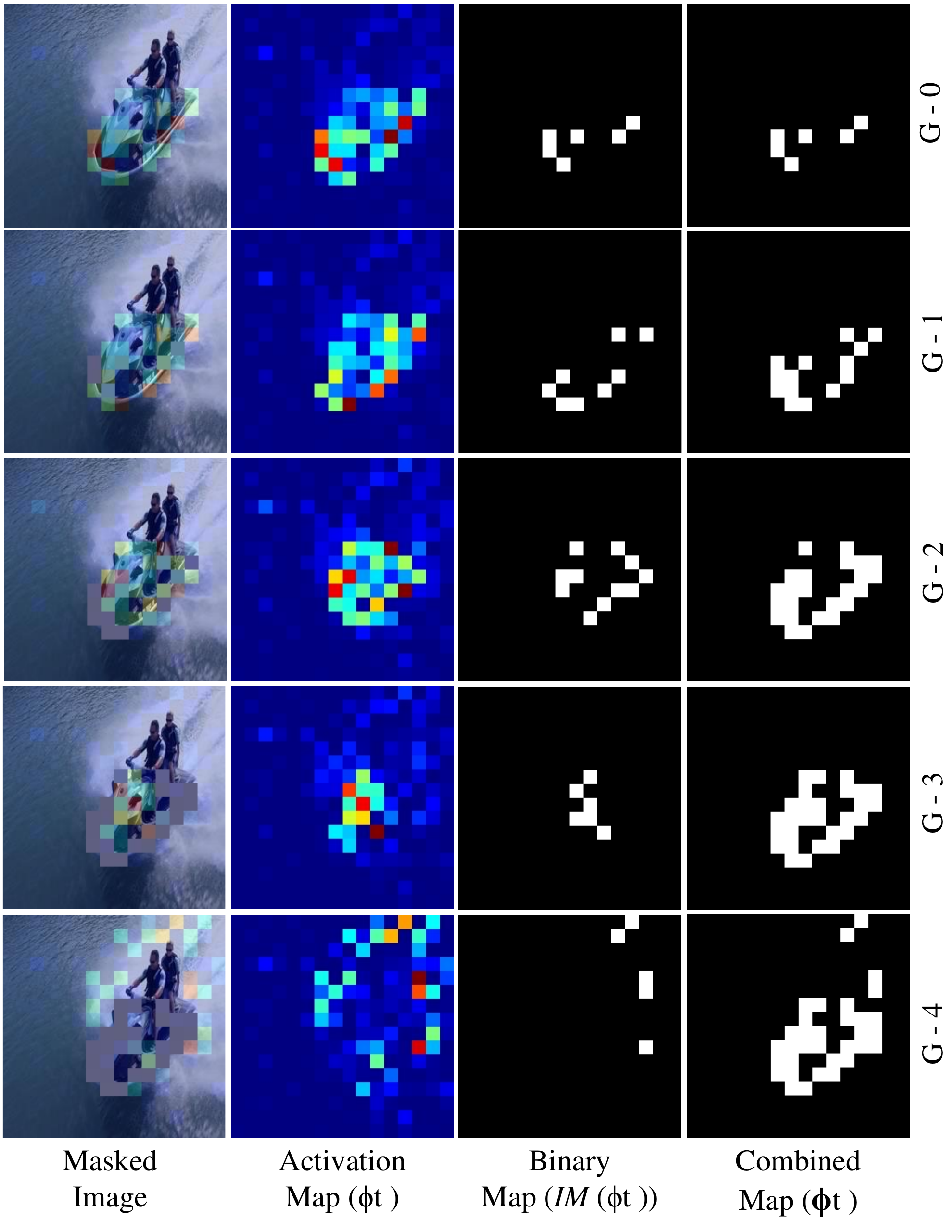}
\end{center}
\vskip -20.0pt \caption{Iterative masking visualization for "Jet ski" category.}
\label{fig:itermask2}
\vskip -20.0pt 
\end{figure*}

\noindent \textbf{Masking for more iteration might produce redundant activation:} \\
Fig. \ref{fig:itermask1} present iterative masking visualization for $G$ going from 0 to 4.
From Fig. \ref{fig:itermask1}, we can observe that the initial GradCAM~\cite{selvaraju2017grad} activation for the Jet ski category is towards the most discriminative parts. Also, the activation for the person category is negligible due to the good region-text alignment property of pre-trained vision-language model. However, as the masking iteration increases after $G$=3, we can observe that the water regions around Jet ski are starting to activate. Hence, more steps might completely mask the object and will start producing redundant activation. Quantitatively we observed $G$-3 produces optimal pseudo-labels, as discussed in section 4.3.

\begin{figure*}
\begin{center}
\includegraphics[width=0.82\linewidth]{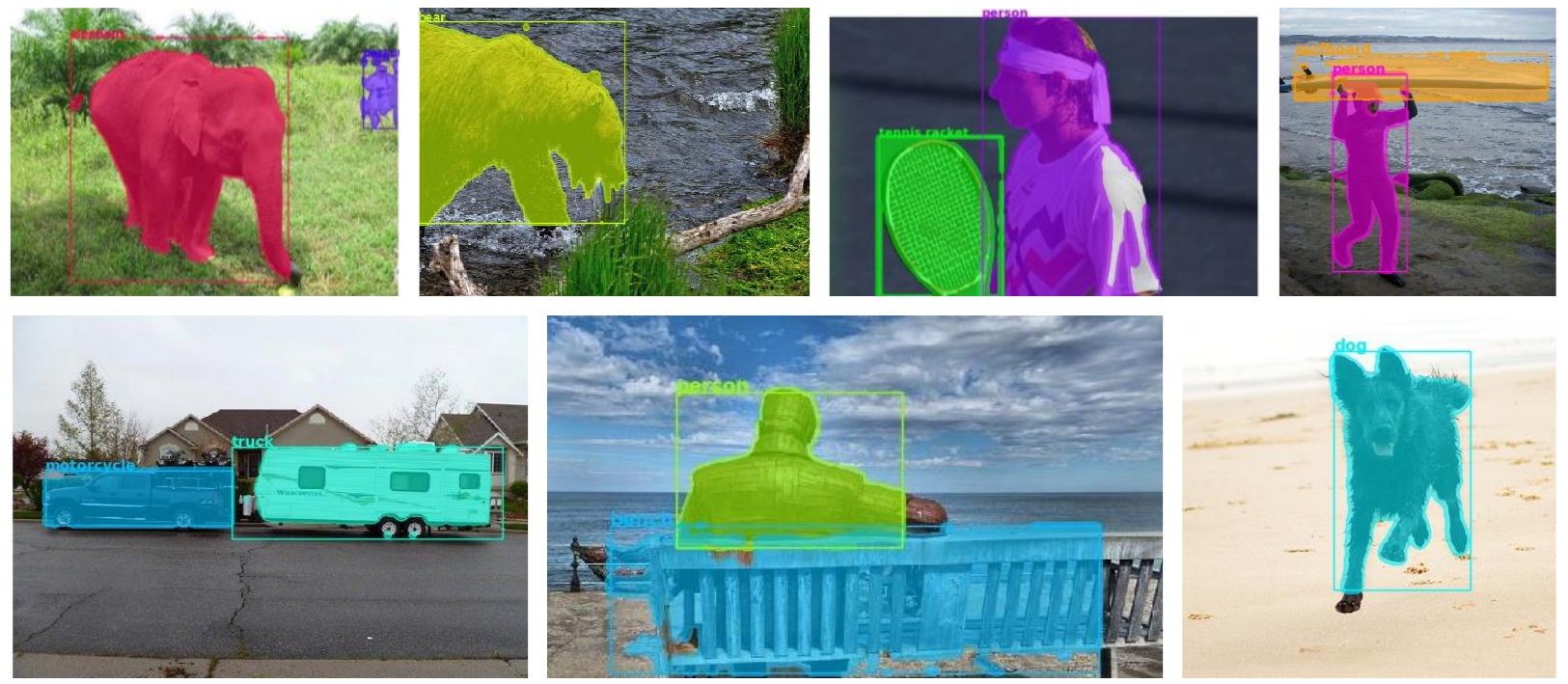}
\end{center}
\vskip -18.0pt \caption{Visualization of pseudo-mask generated for COCO dataset \cite{lin2014microsoft} using our pipeline. Note that, the generated box-level and pixel-level annotations are noisy (incomplete mask and less accurate bounding box).}
\label{fig:sup_graph_vis}
\vskip -12.0pt
\end{figure*}

\begin{figure*}
\begin{center}
\includegraphics[width=0.82\linewidth]{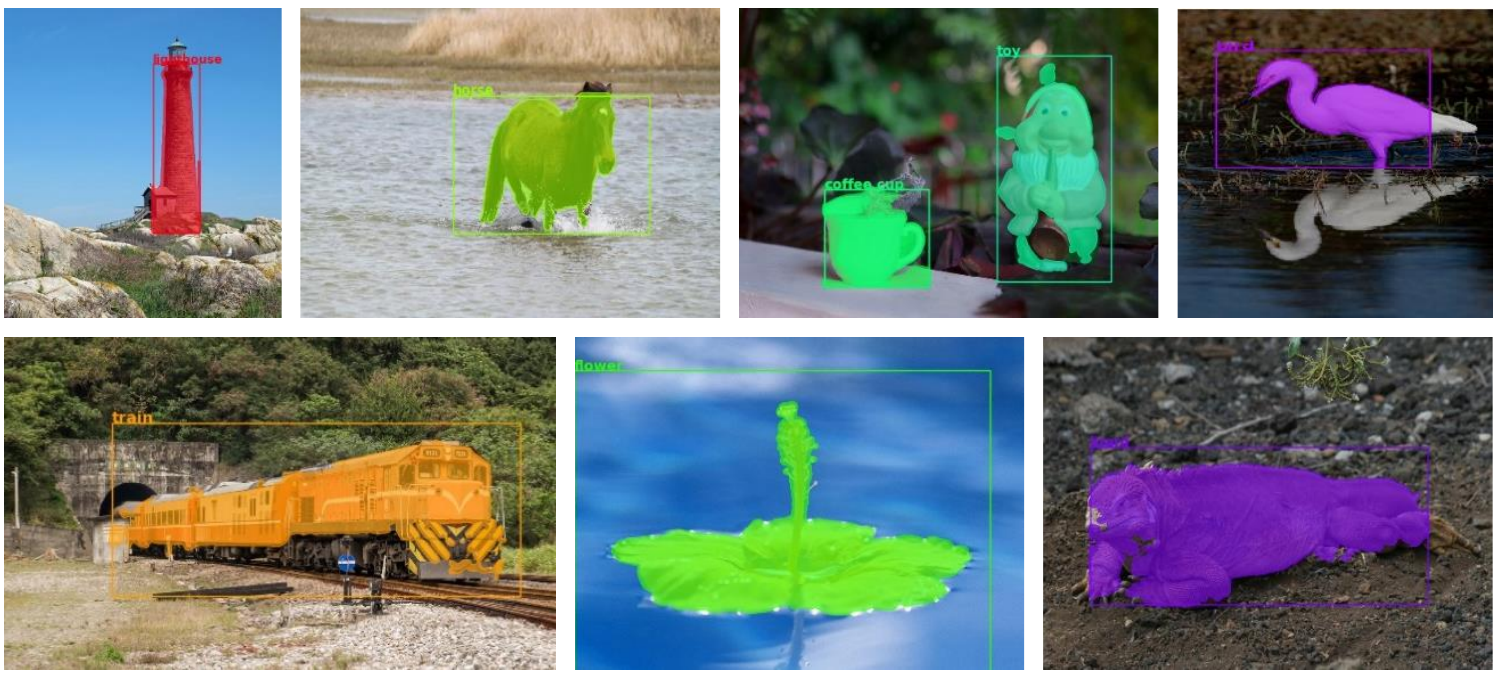}
\end{center}
\vskip -18.0pt \caption{Visualization of pseudo-mask generated for Open Images  \cite{kuznetsova2020open} dataset using our pipeline. Note that, the generated box-level and pixel-level annotations are noisy (incomplete mask and less accurate bounding box).}
\label{fig:sup_graph_vis}
\vskip -12.0pt
\end{figure*}

\begin{figure*}
\begin{center}
\includegraphics[width=0.96\linewidth]{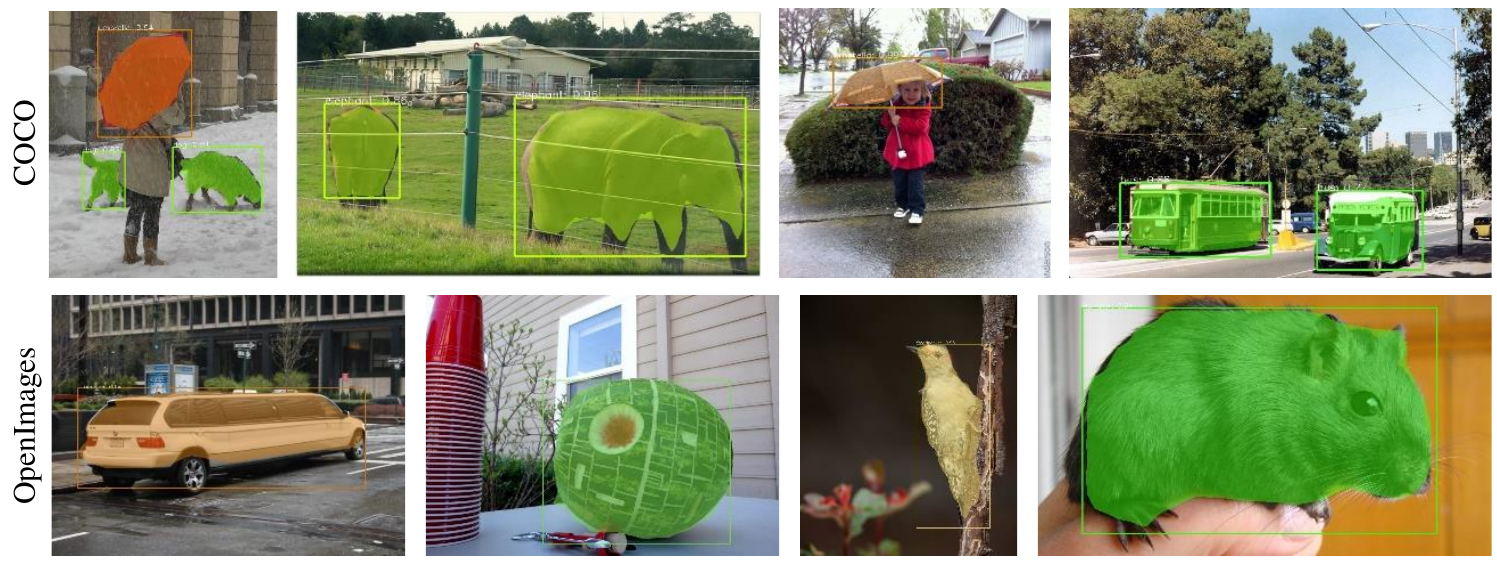}
\end{center}
\vskip -18.0pt \caption{Visualization of Mask-RCNN \cite{he2017mask} predictions trained on pseudo-masks generated on COCO and Open Images - top and bottom row, respectively. Mask-RCNN training helps the model learn to filter the noise present in the pseudo-mask producing better-quality (complete mask and tight bounding box) predictions.}
\label{fig:sup_graph_vis}
\vskip -14.0pt
\end{figure*}

\end{document}